\documentclass[10pt,twocolumn,letterpaper]{article}

\usepackage[pagenumbers]{cvpr} 

\usepackage{graphicx}
\usepackage{amsmath}
\usepackage{amssymb}
\usepackage{booktabs}
\usepackage{times}
\usepackage{epsfig}
\usepackage{bbm}
\usepackage{booktabs}
\usepackage{bm}
\usepackage[usenames,dvipsnames,table]{xcolor}

\usepackage[pagebackref,breaklinks,colorlinks]{hyperref}

\usepackage[capitalize]{cleveref}
\crefname{section}{Sec.}{Secs.}
\Crefname{section}{Section}{Sections}
\Crefname{table}{Table}{Tables}
\crefname{table}{Tab.}{Tabs.}

\def\confName{CVPR}
\def\confYear{2023}

\newcommand{\methodname}{FrustumFormer}
\begin{document}

\title{FrustumFormer: Adaptive Instance-aware Resampling for Multi-view 3D Detection}

\author{
  Yuqi Wang$^{1,2}$\quad 
  Yuntao Chen$^{3}$\quad 
  Zhaoxiang Zhang$^{1,2,3}$ \quad
  \\ 
  \\
$^1$ CRIPAC, Institute of Automation, Chinese Academy of Sciences (CASIA)\\
$^2$ School of Artificial Intelligence, University of Chinese Academy of Sciences (UCAS) \\
$^3$ Centre for Artificial Intelligence and Robotics, HKISI\_CAS \\
  \tt\small{\{wangyuqi2020,zhaoxiang.zhang\}@ia.ac.cn}  \quad \tt\small{chenyuntao08@gmail.com} \\
}
\maketitle

\begin{abstract}
   The transformation of features from 2D perspective space to 3D space is essential to multi-view 3D object detection. 
Recent approaches mainly focus on the design of view transformation, either pixel-wisely lifting perspective view features into 3D space with estimated depth or grid-wisely constructing BEV features via 3D projection, treating all pixels or grids equally.
However, choosing what to transform is also important but has rarely been discussed before.
The pixels of a moving car are more informative than the pixels of the sky.
To fully utilize the information contained in images, the view transformation should be able to adapt to different image regions according to their contents. 
In this paper, we propose a novel framework named \textbf{\methodname{}}, which pays more attention to the features in instance regions via adaptive instance-aware resampling. 
Specifically, the model obtains instance frustums on the bird's eye view by leveraging image view object proposals.
An adaptive occupancy mask within the instance frustum is learned to refine the instance location.
Moreover, the temporal frustum intersection could further reduce the localization uncertainty of objects.
Comprehensive experiments on the nuScenes dataset demonstrate the effectiveness of \methodname{}, and we achieve a new state-of-the-art performance on the benchmark.
Codes and models will be made available at \url{https://github.com/Robertwyq/Frustum}.
\end{abstract}

\begin{figure*}
  \centering
  \begin{subfigure}{0.33\linewidth}
    \includegraphics[width=1.0\linewidth]{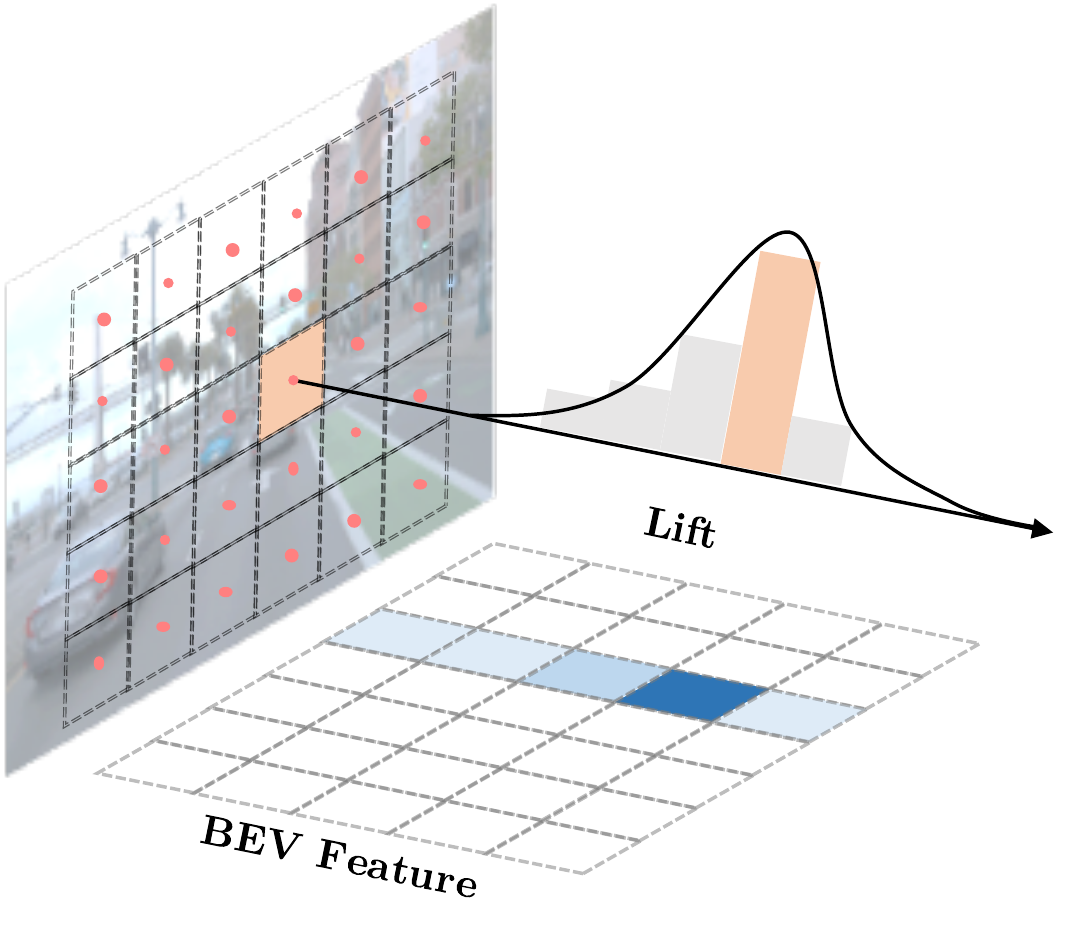}
    \caption{Grid Sampling in Image.}
    \label{fig:imagesample}
  \end{subfigure}
  \hfill
  \begin{subfigure}{0.33\linewidth}
    \includegraphics[width=1.0\linewidth]{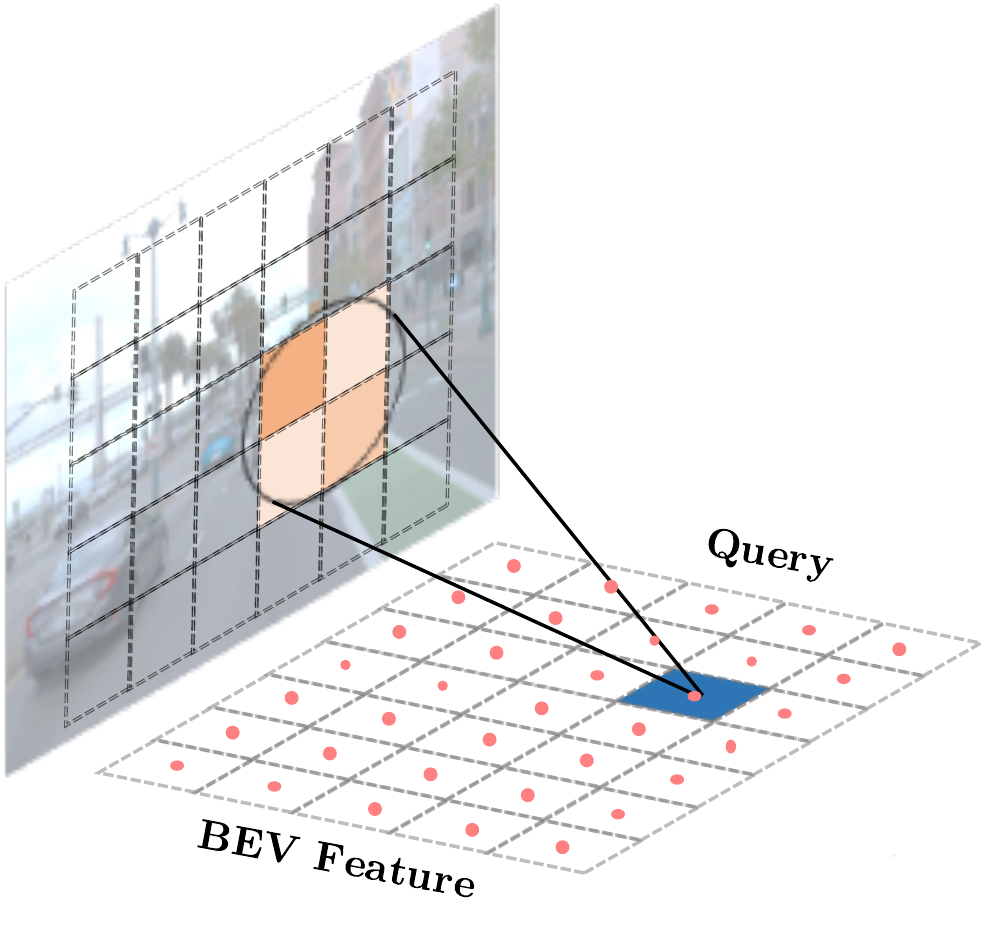}
    \caption{Grid Sampling in BEV.}
    \label{fig:bevsample}
  \end{subfigure}
  \begin{subfigure}{0.33\linewidth}
    \includegraphics[width=1.0\linewidth]{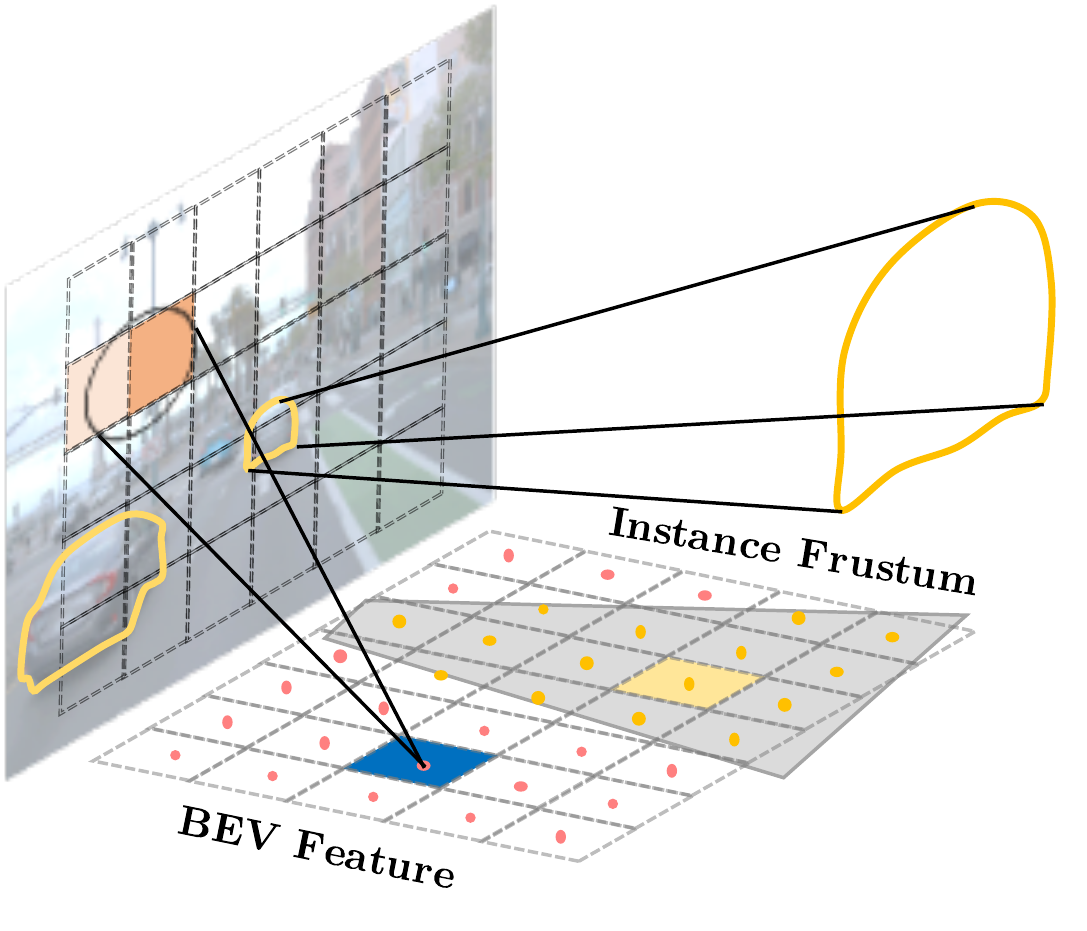}
    \caption{Instance-aware Sampling in Frustum.}
    \label{fig:airsample}
  \end{subfigure}
    \caption{\textbf{Comparison of different sampling strategies for the feature transformation from image view to bird's eye view.} (a) represents the sampling in image view and lift features~\cite{huang2021bevdet} to BEV plane with pixel-wise depth estimation. (b) shows the grid sampling in BEV and queries back~\cite{li2022bevformer} to obtain image features via cross-attention. (c) illustrates our proposed instance-aware sampling strategy in the frustum, which adapts to the view content by focusing more attention on instance regions. This approach is designed to enhance the learning of instance-aware BEV features.}
  \label{fig:frustum}
  \vspace{-10pt}
\end{figure*}
\section{Introduction}
\label{sec:intro}
Perception in 3D space has gained increasing attention in both academia and industry. Despite the success of LiDAR-based methods~\cite{zhou2018voxelnet, lang2019pointpillars,shi2019pointrcnn, yin2021center}, camera-based 3D object detection~\cite{zhang2021objects,wang2021fcos3d,wang2022detr3d,li2022bevformer} has earned a wide audience, due to the low cost for deployment and advantages for long-range detection.
Recently, multi-view 3D detection in Bird's-Eye-View (BEV) has made fast progresses. 
Due to the unified representation in 3D space, multi-view features and temporal information can be fused conveniently, which leads to significant performance improvement over monocular methods~\cite{chen2016monocular,weng2019monocular, wang2021fcos3d,park2021pseudo}.

Transforming perspective view features into the bird's-eye view is the key to the success of modern BEV 3D detectors~\cite{huang2021bevdet,li2022bevformer,liu2022petr,li2022bevdepth}.
As shown in \cref{fig:frustum}, we categorize the existing methods into lifting-based ones like LSS~\cite{philion2020lift} and BEVDet~\cite{huang2021bevdet} and query-based ones like BEVFormer~\cite{li2022bevformer} and Ego3RT~\cite{lu2022learning}.
However, these methods mainly focus on the design of view transformation strategies while overlooking the significance of choosing the right features to transform during view transformation.
Regions containing objects like vehicles and pedestrians are apparently more informative than the empty background like sky and ground.
But all previous methods treat them with equal importance.
We suggest that the view transformation should be adaptive with respect to the image content. 
Therefore, we propose \emph{Adaptive Instance-aware Resampling (AIR)}, an instance-aware view transformation, as shown in \cref{fig:airsample}. 
The core idea of AIR is to reduce instance localization uncertainty by focusing on a selective part of BEV queries.
Localizing instance regions is difficult directly on the BEV plane but relatively easy in the image view.
Therefore, the \emph{instance frustum}, lifting from instance proposals in image views, gives geometrical hints of the possible locations of objects in the 3D space.
Though the instance frustum has provided initial prior locations, it is still a large uncertain area. 
We propose an \emph{occupancy mask predictor} and a \emph{temporal frustum fusion module} to further reduce the localization uncertainty.
Our model learns an occupancy mask for frustum queries on the BEV plane, predicting the possibility that a region might contain objects.
We also fuse instance frustums across different time steps, where the intersection area poses geometric constraints for actual locations of objects.

We propose a novel framework called \emph{\methodname{}} based on the insights mentioned previously, which effectively enhances the learning of instance-aware BEV features via Adaptive Instance-aware Resampling. 
\methodname{} utilizes the instance frustum to establish the connection between perspective and bird's eye view regions, which contains two key designs:
(1) A frustum encoder that enhances instance-aware features via adaptive instance-aware resampling.
(2) A temporal frustum fusion module that aggregates historical instance frustum features for accurate localization and velocity prediction.
In conclusion, the contributions of this work are as follows:
\begin{itemize}
    \item We propose \emph{\methodname{}}, a novel framework that exploits the geometric constraints behind perspective view and birds' eye view by instance frustum.
    \item We propose that choosing what to transform is also important during view transformation. The view transformation should adapt to the view content. Instance regions should gain more attention rather than be treated equally. Therefore, we design \emph{Adaptive Instance-aware Resampling (AIR)} to focus more on the instance regions, leveraging sparse instance queries to enhance the learning of instance-aware BEV features.
    \item We evaluate the proposed \emph{\methodname{}} on the nuScenes dataset. We achieve improved performance compared to prior arts. \emph{\methodname{}} achieves 58.9 NDS and 51.6 mAP on nuScenes test set without bells and whistles.
\end{itemize}

\begin{figure*}
  \centering
  \begin{subfigure}{0.74\linewidth}
    \includegraphics[width=1.0\linewidth]{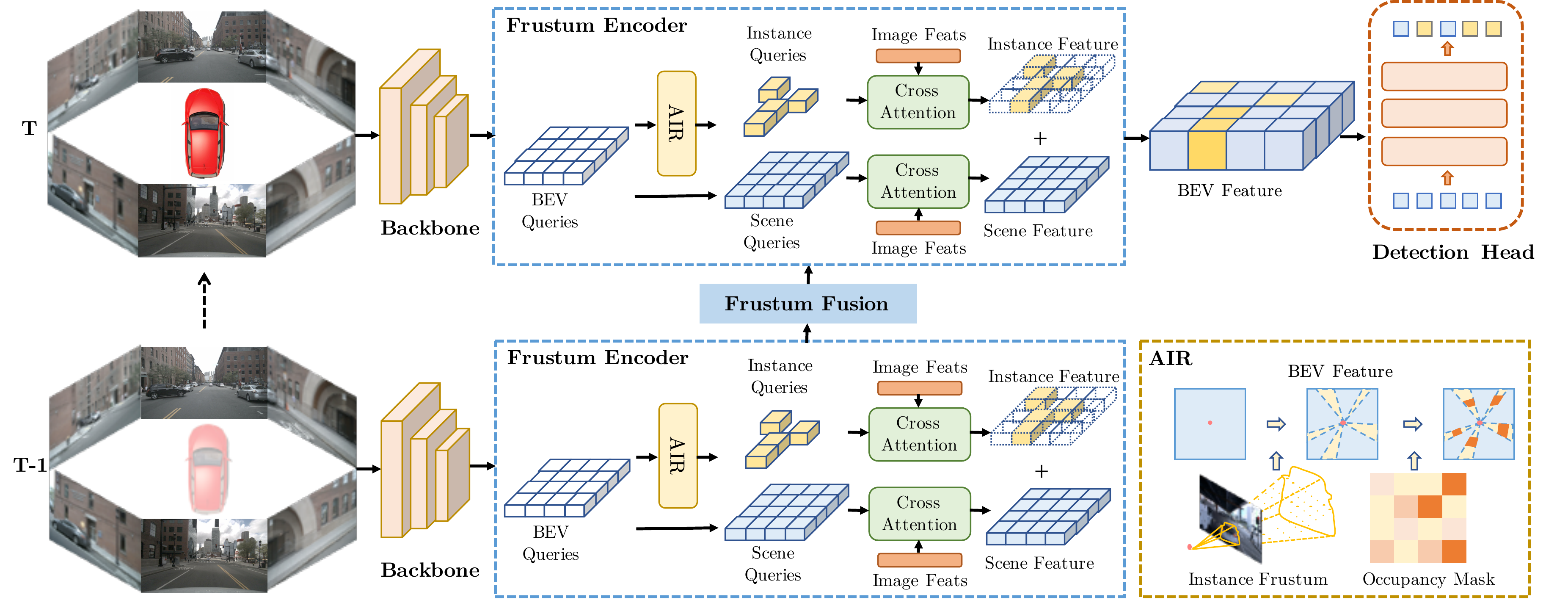}
    \caption{Overall architecture of \methodname{}.}
    \label{fig:architecture}
  \end{subfigure}
  \hfill
  \begin{subfigure}{0.24\linewidth}
    \includegraphics[width=1.0\linewidth]{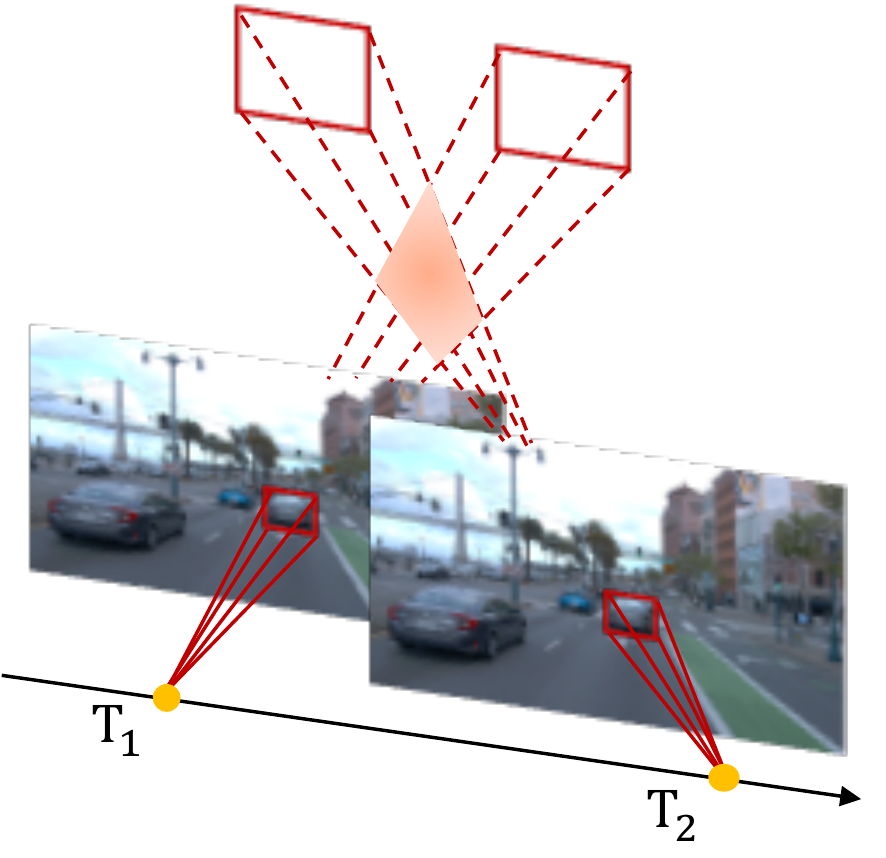}
    \caption{Temporal Frustum Fusion.}
    \label{fig:ff}
  \end{subfigure}
  \caption{\textbf{Illustration of our proposed \methodname{}.} (a) shows the overall pipeline. The image backbone first extracts the multi-view image features. These features are transformed into a unified BEV feature by the frustum encoder, which integrates temporal information from frustum fusion. Then the detection head decodes the BEV feature to the final outputs. Adaptive Instance-aware Resampling (AIR) is utilized to adaptively adjust the sampling area and select instance queries according to the view content. It consists of instance frustum query generation and frustum occupancy mask prediction. (b) illustrates the hints for object locations during temporal frustum fusion.}
  \label{fig:FrustumFormer}
  \vspace{-10pt}
\end{figure*}
\section{Related Work}
\label{sec:related}
\subsection{Frustum-based 3D Object Detection}
Frustum indicates the possible locations of 3D objects in a 3D space by projecting 2D interested regions. The frustum is commonly used to aid fusion~\cite{qi2018frustum,wang2019frustum,zhang2021faraway,paigwar2021frustum,erccelik2021temp} in 3D object detection when RGB images and LiDAR data are available. 
Frustum PointNets~\cite{qi2018frustum} takes advantage of mature 2D object detectors and performs 3D object instance segmentation within the trimmed 3D frustums.
Frustum Fusion~\cite{negahbani2021frustum} leverages the intersection volume of the two frustums induced by the 2D detection on stereo images. To deal with LiDAR sparsity, Faraway-Frustum~\cite{zhang2021faraway} proposes a novel fusion strategy for detecting faraway objects.
In this paper, we introduce the idea of frustum into camera-only 3D detection for enhancing instance-aware BEV features.

\subsection{Multi-view 3D Object Detection}
Multi-view 3D object detection aims to predict the 3D bounding boxes and categories of the objects with multi-view images as input.
Current methods can be divided into two schemes: \emph{lifting 2D to 3D} and \emph{Querying 2D from 3D}.

\noindent \textbf{Lifting 2D to 3D.} 
Following the spirit of LSS~\cite{philion2020lift}, BEVDet~\cite{huang2021bevdet} lifts multi-view 2D image features into a depth-aware frustum and splats into a unified bird's-eye-view (BEV) representation and applies to the detection task. BEVDepth~\cite{li2022bevdepth} utilizes LiDAR points as supervision to learn reliable depth estimation. BEVDet4D~\cite{huang2022bevdet4d} incorporates the temporal information and extends the BEVDet to the spatial-temporal 4D working space. Recently, STS~\cite{wang2022sts}, BEVStereo~\cite{li2022bevstereo} and SOLOFusion~\cite{park2022time} further attempt to improve the depth learning by 
combining temporal geometric constraints.
Overall, these works demonstrate the importance of incorporating depth and temporal information in improving detection performance.

\noindent \textbf{Querying 2D from 3D.}
Following DETR~\cite{carion2020end}, DETR3D~\cite{wang2022detr3d} predicts learnable queries in 3D space and projects back to query the corresponding 2D image features. PETR~\cite{liu2022petr,liu2022petrv2} proposes to query directly with 3D position-aware features, which are generated by encoding the 3D position embedding into 2D image features.
Ego3RT~\cite{lu2022learning} introduces the polarized grid of dense imaginary eyes and sends rays backward to 2D visual representation. BEVFormer~\cite{li2022bevformer} learns spatiotemporal BEV features via deformable attention, which explicitly constructs the BEV grid samples in 3D space and queries back to aggregate multi-view image features. PolarFormer~\cite{jiang2022polarformer} further generates polar queries in the Polar coordinate and encodes BEV features in Polar space.

\section{Method}
In multi-view 3D object detection task, $N$ monocular views of images $I = \{{I_i \in \mathbb{R}^{3\times H\times W}}\}_{i=1}^N$, together with camera intrinsics $K = \{K_i \in \mathbb{R}^{3 \times 3} \}_{i=1}^N$ and camera extrinsics $T = \{T_i \in \mathbb{R}^{4 \times 4} \}_{i=1}^N$ are given. 
The objective of the model is to output the 3D attributes (locations, size, and velocity) and the corresponding category of objects contained in multi-view images.

As shown in~\cref{fig:architecture}, \methodname{} mainly focuses on the feature transformation process and is composed of four components: an image backbone, a frustum encoder, a frustum fusion module, and a detection head.
The image backbone first extracts multi-scale image features from multi-view images.  Aiding by the frustum fusion module, the image features transform into a unified BEV feature via the frustum encoder. 
Finally, a query-based detection head decodes the BEV feature to the outputs of the detection task.

\subsection{Frustum Encoder}
Frustum encoder transforms the multi-scale multi-view image features $\mathbf{F}$ to a unified BEV feature $\mathbf{B}$.
Instead of treating all regions equally during the feature view transformation, our frustum encoder adaptively transforms the image features according to the view content.
As shown in \cref{fig:architecture}, we use two types of BEV queries to construct the final BEV features, the scene queries $ \mathbf{Q}_s$ and the instance queries $\mathbf{Q}_i$.
Scene queries(blue grids) are dense and generated from regular grids, while instance queries(yellow grids) are sparse and generated from irregular instance frustum.
The learning process of the scene query is similar to BEVFormer~\cite{li2022bevformer}. 
However, the instance queries are learned inside instance regions, and the learned instance feature are further combined with the scene feature to form the final instance-aware BEV features. 
Specifically, the selection of instance regions is made via \emph{adaptive instance-aware resampling}, which consists of \emph{(1) instance frustum query generation} and \emph{(2) frustum occupancy mask prediction}. 
Finally, the instance feature is learned by \emph{(3) instance frustum cross-attention} computed in selected instance regions.
We will introduce these three parts in the following.

\noindent \textbf{Instance Frustum Query Generation.}
This section introduces the query generation for a single instance frustum $\mathbf{Q}_f$, which is a subset of instance queries $\mathbf{Q}_i$.
The core insight is to leverage the instance mask from perspective views and select the corresponding region on the BEV plane.
Following the query-based~\cite{li2022bevformer,jiang2022polarformer} view transformation, we define a group of grid-shape learnable parameters $\mathbf{Q}_i \in \mathbb{R}^{H \times W \times C}$ as the instance queries. 
$H$, $W$ are the spatial shape of BEV queries, and $C$ is the channel dimension.
We first generate sampling points $\{\mathbf{p}_i^k=(x_i,y_i,z_k), i \in H\times W, k \in K\}$ corresponding to a single BEV query $\mathbf{Q}^{p_i}$ at grid region center $\mathbf{p}_i = (x_i,y_i)$, and then project these points to different image views. 
$K$ is the number of sampling points in the vertical direction of pillars.
The projection between sampling points $\mathbf{p}_i^k$ and its corresponding 2D reference point $(u_{ij}^k,v_{ij}^k)$ on the $j$-th image view is formulated as:
\begin{align}
    \pi_j(\mathbf{p}_i^k)& = (u_{ij}^k,v_{ij}^k) \\
    d_{ij}^k\cdot[u_{ij}^k,v_{ij}^k,1]^T & = \mathbf{T}_j \cdot [x_i^k, y_i^k, z_i^k, 1]^T
\end{align}
where $\pi_j(\mathbf{p}_i^k)$ denotes the projection of the $k$-th sampling point at location $\mathbf{p}_i$ on the $j$-th camera view. $\mathbf{T}_j \in \mathbb{R}^{3 \times 4}$ is the projection matrix of the $j$-th camera. $x_i^k,y_i^k,z_i^k$ represents the 3D location of the sampling point in the vehicle frame. $u_{ij}^k,v_{ij}^k$ denotes corresponding 2D reference point on $j$-th view projected from 3D sampling point $\mathbf{p}_i^k$. $d_{ij}^k$ is the depth in the camera frame.

Predicting the instance frustum region directly in bird's eye view is challenging, but detecting the objects in perspective view~\cite{ren2015faster,he2017mask} is relatively mature.
Inspired by this, we take advantage of object masks on the image plane and leverage its geometric clues for the BEV plane. 
The instance frustum queries $\mathbf{Q}_f$ of a specific 2D instance could be defined as all instance BEV queries $\mathbf{Q}_i$ with image plane projection points inside the object mask $S$ in~\cref{eq:instancefrustum}:
\begin{equation}
     \mathbf{Q}_f = \{\mathbf{Q}^{p_i}\in \mathbf{Q}_i | \quad \exists \quad \pi (\mathbf{p}_i^k)\in S\}
    \label{eq:instancefrustum}
\end{equation}
$\mathbf{Q}^{p_i}\in \mathbb{R}^{1\times C}$ is the query located at $\mathbf{p}_i=(x_i,y_i)$. $\mathbf{p}_i^k$ represents the $k$-th sampling points in the pillars at $\mathbf{p}_i$.
$\pi (\mathbf{p}_i^k)$ denotes the projection points of $\mathbf{p}_i^k$ to the image plane.

\noindent \textbf{Frustum Occupancy Mask Prediction.}
Although the instance frustum provides potential locations for objects, its corresponding area on the BEV plane is often large due to depth uncertainty.
To reduce this localization uncertainty, we propose to predict an occupancy mask for all frustums.
Specifically, given the union of all instance frustum queries $\cup\mathbf{Q}_f$, we design a \textit{OccMask} module to predict a binary occupancy mask $\mathbf{O}_{bev}\in \mathbb{R}^{H\times W\times1}$ on the BEV plane for all instance frustum queries in a single shot. 
The occupancy mask reflects the probability of a grid-wise region containing the objects, and is computed by \cref{eq:occupancy}:
\begin{equation}
    \mathbf{O}_{bev} = \textit{OccMask}(\cup\mathbf{Q}_f)
    \label{eq:occupancy}
\end{equation}
\textit{OccMask} is a learned module composed of 2D convolutions. The supervision comes from the projection of the ground truth 3D bounding boxes on the BEV plane.
We use the focal loss~\cite{lin2017focal} for learning occupancy mask in \cref{eq:loss}:
\begin{equation}
    \mathcal{L}_{m} = \textit{Focal Loss}(\mathbf{O}_{bev},\mathbf{\Omega})
    \label{eq:loss}
\end{equation}
where $\mathbf{\Omega}$ represents the projection mask of the 3D bounding boxes onto the BEV plane. We select the minimum projecting bounding box on the BEV plane as the supervisory signal, which is made up of the outermost corners of the objects, considering their rotation.
To further refine our predictions, the supervision signal for $\mathbf{O}_{bev}$ is added to each layer of the frustum encoder for iterative refinement, alongside BEV instance feature learning. In each layer, we predict the current occupancy mask using the last output instance frustum query from the previous layer. This approach enables the sampling areas to adapt to the previous layer output.

\noindent \textbf{Instance Frustum Cross-Attention.} 
Instance Frustum Cross-Attention (IFCA) is designed for the feature interaction between instance queries $\mathbf{Q}_{i}$ and image view features $\mathbf{F}$.
The instance queries $\mathbf{Q}_{i}$ is selected by \cref{eq:R}: 
\begin{equation}
    \mathbf{Q}_{i} = \{\mathbf{Q}^{p_i}\in \cup\mathbf{Q}_f | \mathbf{O}_{bev}(i)=1\}
    \label{eq:R}
\end{equation}
Instance queries are selected from the instance frustum queries $\mathbf{Q}_f$. $\mathbf{O}_{bev}(i)$ denotes the occupancy value at position $\mathbf{p}_i$ on the BEV plane.
For each query $\mathbf{Q}^{p_i}$ in $\mathbf{Q}_f$, if $\mathbf{O}_{bev}(i)$ predicts the occupancy value is 1, then the query $\mathbf{Q}^{p_i}$ is marked as instance query.
The process of instance frustum cross-attention (IFCA) can be formulated as:
\begin{equation}
    \textit{IFCA}(\mathbf{Q}_{i}^{p_i},\mathbf{F}_j) = \frac{1}{|v|}\sum_{j\in v}\sum_{m=1}^M \textit{DA}(\mathbf{Q}_{i}^{p_i},\pi_j(\mathbf{p}_i^m),\mathbf{F}_j)
\end{equation}
where $\mathbf{Q}_{i}^{p_i}$ is an instance query at location $\mathbf{p}_i$, $\pi_j(\mathbf{p}_i^m)$ is the projection to get the $m$-th 2D reference point on the $j$-th camera view. $M$ is the total number of sampling points for an instance query. $\mathbf{F}_j$ is the image features of the $j$-th camera view. \textit{DA} represents deformable attention. $v$ is the set of image views for which the 2D reference point can fall on. $|v|$ represents the number of views.

\subsection{Frustum Fusion Module}
Temporal information is essential for camera-based 3D object detection, especially in inferring the motion state of objects and recognizing objects under heavy occlusions.
Beyond learning occupancy mask on the BEV plane, another solution for eliminating the location uncertainty in the instance frustum is to fuse the temporal information.

\noindent \textbf{Temporal Frustum Intersection.}
As shown in~\cref{fig:ff}, the intersection area of the instance frustum at different timestamps leaves hints for the accurate location of 3D objects.
Inspired by this, we constrain the query interaction within instance frustum regions, implicitly learning features from interaction areas.
Given instance frustum queries $\mathbf{Q}_f$ at current timestamp $t$ and history instance frustum queries $\mathbf{H}_f$ preserved at timestamp $t'$. 
For a query $\cup\mathbf{Q}_f^{p_i}$ at position $\mathbf{p}_i$, we use the information from ego-motion $(\Delta x, \Delta y, \Delta \theta)$ to compute the corresponding position $\mathbf{p'}_i$ at timestamp $t'$. 
The cross-attention for query $\mathbf{Q}_f^{p_i}$ only compute the history queries around position $\mathbf{p'}_i$ of $\mathbf{H}_f$.
Following~\cite{li2022bevformer}, we adopt a sequential RNN-like~\cite{cho2014properties} way of fusing the historical instance frustum queries. This approach enables the aggregation of long-range hints for the intersection area.

\noindent \textbf{Temporal Frustum Cross-Attention.}
Temporal Frustum Cross-Attention (TFCA) aggregates the information of history instance frustum queries $\mathbf{H}_f$ into the current instance frustum queries $\mathbf{Q}_f$.
Since the objects might be movable in the scene, causing the misalignment if only computing the query at $\mathbf{p'}_i$. Deformable attention~\cite{zhu2020deformable} is utilized to reduce the influence of object movement.
The process of temporal frustum cross-attention (TFCA) can be formulated as follows:
\begin{equation}
    \textit{TFCA}(\mathbf{Q}_f^{p_i},\mathbf{H}_f)=\sum_{m=1}^M \textit{DA}(\mathbf{Q}_f^{p_i},\mathbf{p'}_i^m,\mathbf{H}_f)
\end{equation}
where $\mathbf{Q}_f^{p_i}$ denotes the instance frustum query located at $\mathbf{p}_i = (x_i,y_i)$. $\mathbf{H}_f$ represents the history instance frustum query. $\mathbf{p'}_i$ is the aligned position by ego-motion. For each query at location $\mathbf{p'}_i$, we sample $M$ points $\mathbf{p'}_i^m$ to query the history instance frustum feature. \textit{DA} represents deformable attention.
\section{Experiment}

\subsection{Datasets}
\noindent \textbf{nuScenes dataset}~\cite{caesar2020nuscenes}. 
The nuScenes dataset provides 1000 sequences of different scenes collected in Boston and Singapore. These sequences are officially split into 700/150/150 ones for training, validation, and testing. Each sequence is roughly about 20s duration, and the key samples are annotated at 2Hz, contributing to a total of 1.4M objects bounding boxes. Each sample consists of RGB images from 6 cameras covering the 360-degree horizontal FOV: front, front left, front right, back, back left, and back right. The image resolution is 1600$\times$900 pixels in all views.
10 classes are annotated for the object-detecting task: car, truck, bus, trailer, construction vehicle, pedestrian, motorcycle, bicycle, barrier, and traffic cone.

\noindent \textbf{Evaluation metrics.} 
For the official evaluation protocol in the nuScenes dataset, the metrics include mean Average Precision (mAP) and a set of True Positive (TP) metrics, which contains the average translation error (ATE), average scale error (ASE), average orientation error (AOE), average velocity error (AVE), and average attribute error (AAE). Finally, the nuScenes detection score (NDS) is defined to consider the above metrics.

\subsection{Experimental Settings}
\noindent \textbf{Implementation Details.}
Following previous methods~\cite{wang2021fcos3d,wang2022detr3d,li2022bevformer}, we utilize two types of backbone: ResNet101-DCN~\cite{he2016deep,dai2017deformable} that initialized from FCOS3D~\cite{wang2021fcos3d}, and VoVnet-99~\cite{lee2019energy} that initialized from DD3D~\cite{park2021pseudo}.
We utilize the output multi-scale features from FPN~\cite{lin2017feature} with sizes of 1/8, 1/16, 1/32, and 1/64 and a feature dimension of 256. The frustum encoder includes 6 layers and is implemented based on BEVFormer~\cite{li2022bevformer}. The default size of BEV queries is 200x200, and the perception ranges are [-51.2m, 51.2m] for the $X$ and $Y$ axis and [-3m, 5m] for the $Z$ axis. We sample K=8 points for each pillar-like region of the BEV query. We adopt learnable position embedding for BEV queries. 
There are two types of queries in frustum encoder: sparse instance queries and dense scene queries. we followed BEVformer~\cite{li2022bevformer} to extract the scene feature by scene queries. The instance feature is extracted by instance queries.
For the 2D instance proposals, we utilized the Mask R-CNN~\cite{he2017mask} pre-trained on the nuImages~\cite{caesar2020nuscenes}.
We use the output bounding boxes to generate object mask regions, and the score threshold is set to 0.5.
The loss weight for $\mathcal{L}_m$ is set to 5.
The frustum fusion module uses a temporal window size of 8 and randomly samples 4 key-frames during training.
We adopted a query-based detection head to decode the BEV features, the same in~\cite{li2022bevformer}. The number of the object query is set to 900 and has 3 groups of queries~\cite{chen2022group} during training.

\noindent \textbf{Training.}
We train the model on 8 NVIDIA A100 GPUs with batch size 1 per GPU. 
We train our model with AdamW~\cite{loshchilov2018fixing} optimizer for 24 epochs, an initial learning rate of $2\times 10^{-4}$ with a cosine annealing schedule.
The input of the images is cropped to 1600 $\times$ 640. We adopt data augmentations like image scaling, flipping, color distortion, and GridMask~\cite{chen2020gridmask}. 
For the ablation study, we train the model with a total batch size of 8 for 24 epochs without data augmentation. We use the ResNet-50~\cite{he2016deep} as the backbone. The image resolution is resized at a scale of 0.8, which is 1280 $\times$ 512.

\noindent \textbf{Inference.}
During inference, the previous BEV features are saved and used for the next, corresponding to the infinite temporal window of a sequence. This online inference strategy is time-efficient.
Since we adopted three groups of queries during training, only one group is utilized at inference time.
We do not adopt model-agnostic tricks such as model ensemble and test-time augmentation when evaluating our model on both \emph{val} and \emph{test} sets.

\begin{table*}[ht]
\centering
\tiny
\resizebox{\textwidth}{!}{
\begin{tabular}{l|c|c|c|c|c|ccc@{\hspace{1.0\tabcolsep}}c@{\hspace{1.0\tabcolsep}}c@{\hspace{1.0\tabcolsep}}c@{\hspace{1.0\tabcolsep}}c} 
\toprule
\textbf{Methods} & \textbf{Backbone}&\textbf{CBGS} &\textbf{LiDAR}  & \textbf{mAP}$\uparrow$  &\textbf{NDS}$\uparrow$  & \textbf{mATE}$\downarrow$ & \textbf{mASE}$\downarrow$   &\textbf{mAOE}$\downarrow$   &\textbf{mAVE}$\downarrow$   &\textbf{mAAE}$\downarrow$  \\
\midrule
FCOS3D$\ddag$~\cite{wang2021fcos3d} & R101$^\dag$& & & 0.358 & 0.428 & 0.690 & 0.249 & 0.452 & 1.434 & \textbf{0.124} \\ 
PGD~\cite{wang2022probabilistic} & R101$^\dag$& & & 0.386 & 0.448 & 0.626 & \textbf{0.245} & 0.451 & 1.509 & 0.127 \\
BEVFormer~\cite{li2022bevformer} & R101$^\dag$& && 0.445 & 0.535 & 0.631 & 0.257 & 0.405 & 0.435 & 0.143 \\
PolarFormer~\cite{jiang2022polarformer} & R101$^\dag$& && 0.457 & 0.543 & 0.612 & 0.257 & \textbf{0.392} & 0.467 & 0.129 \\
\rowcolor[gray]{.9} 
\methodname{} & R101$^\dag$& & &\textbf{0.478}  &\textbf{0.561} & \textbf{0.575} & 0.257 & 0.402& \textbf{0.411} &  0.132\\ 
\midrule
DD3D~\cite{park2021pseudo}$\ddag$ & V2-99* && &0.418 & 0.477 & 0.572 & 0.249 & 0.368 & 1.014 & 0.124 \\
DETR3D$\ddag$~\cite{wang2022detr3d} & V2-99* &\checkmark& &0.412 & 0.479 & 0.641 & 0.255 & 0.394 & 0.845 & 0.133 \\
Ego3RT~\cite{lu2022learning} & V2-99* && &0.425 & 0.473 & 0.549 & 0.264 & 0.433 & 1.014 & 0.145 \\
M2BEV~\cite{xie2022m} & X-101 && & 0.429 & 0.474 & 0.583 & 0.254 & 0.376 & 1.053 & 0.190 \\
BEVDet4D$\ddag$~\cite{huang2022bevdet4d} & Swin-B &\checkmark& & 0.451 & 0.569 & \textbf{0.511} & \textbf{0.241} & 0.386 & \textbf{0.301} & 0.121 \\
UVTR~\cite{li2022unifying} & V2-99* && & 0.472 & 0.551 & 0.577 & 0.253 & 0.391 & 0.508 & 0.123 \\
BEVFormer~\cite{li2022bevformer} & V2-99* && &0.481 & 0.569 & 0.582 & 0.256 & 0.375 & 0.378 & 0.126 \\
PolarFormer~\cite{jiang2022polarformer} & V2-99* &&& 0.493 & 0.572 & 0.556 & 0.256 & 0.364 & 0.440 & 0.127 \\
PETRv2~\cite{liu2022petrv2} & V2-99* &&& 0.490 & 0.582 & 0.561 & 0.243 & \textbf{0.361} & 0.343 & \textbf{0.120} \\
\midrule
BEVDepth$\ddag$~\cite{li2022bevdepth} &  V2-99* &\checkmark& \checkmark& 0.503 & 0.600 & 0.445 & 0.245 & 0.378 & 0.320 & 0.126 \\
BEVStereo~\cite{li2022bevstereo} & V2-99* &\checkmark&\checkmark& 0.525 & 0.610 & 0.431 & 0.246 & 0.358 & 0.357 & 0.138 \\
\midrule
\rowcolor[gray]{.9} 
\methodname{} & V2-99*   & && \textbf{0.516} & \textbf{0.589}  & 0.555 &0.249 & 0.372 &0.389  &0.126  \\
\bottomrule
\end{tabular}}
\caption{\textbf{Comparison to state-of-art on the nuScenes \emph{test} set.} * notes that VoVNet-99(V2-99)~\cite{lee2019energy} was pre-trained on the depth estimation task with extra data~\cite{park2021pseudo}. ${}^\dag$Initialized from FCOS3D~\cite{wang2021fcos3d} backbone. $\ddag$ means utilizing test-time augmentation during inference. The commonly used scheme for training is 24 epochs, and \emph{CBGS}~\cite{zhu2019class} would increase the training epochs by nearly $4.5\times$. \emph{LiDAR} means training depth branch utilizing extra modality supervision from LiDAR.}
\label{tab:main_test_set}
\end{table*}

\begin{table*}[ht]
\centering
\tiny
\resizebox{\textwidth}{!}{
\begin{tabular}{l|c|c|c|c|c|ccc@{\hspace{1.0\tabcolsep}}c@{\hspace{1.0\tabcolsep}}c@{\hspace{1.0\tabcolsep}}c@{\hspace{1.0\tabcolsep}}c@{\hspace{1.0\tabcolsep}}c} 

\toprule
\textbf{Methods} & \textbf{Backbone} & \textbf{CBGS} & \textbf{LiDAR} & \textbf{mAP}$\uparrow$  &\textbf{NDS}$\uparrow$  & \textbf{mATE}$\downarrow$ & \textbf{mASE}$\downarrow$   &\textbf{mAOE}$\downarrow$   &\textbf{mAVE}$\downarrow$   &\textbf{mAAE}$\downarrow$  \\
\midrule
FCOS3D~\cite{wang2021fcos3d} & R101$^\dag$ &  & & 0.295 & 0.372 & 0.806 & 0.268 & 0.511 & 1.315 & 0.170 \\ 
DETR3D~\cite{wang2022detr3d} & R101$^\dag$  &\checkmark &  & 0.349 & 0.434 & 0.716 & 0.268 & 0.379 & 0.842 & 0.200 \\
PGD~\cite{wang2022probabilistic} & R101$^\dag$  & & & 0.358 & 0.425 & 0.667 & \textbf{0.264} & 0.435 & 1.276 & \textbf{0.177} \\
PETR~\cite{liu2022petr} & R101$^\dag$ &\checkmark & & 0.370 & 0.442 & 0.711 & 0.267 & 0.383 & 0.865 & 0.201 \\
UVTR~\cite{li2022unifying} & R101$^\dag$  &&  & 0.379 & 0.483 & 0.731 & 0.267 & 0.350 & 0.510 & 0.200 \\
BEVFormer~\cite{li2022bevformer} & R101$^\dag$& &  & 0.416 & 0.517 & 0.673 & 0.274 & 0.372 & 0.394 & 0.198 \\ 
PolarFormer~\cite{jiang2022polarformer} & R101$^\dag$ & &  & 0.432 & 0.528 & 0.648 & 0.270 & \textbf{0.348} & 0.409 & 0.201 \\ 
\midrule
BEVDepth~\cite{li2022bevdepth} & R101 &\checkmark &\checkmark  & 0.412 & 0.535 & 0.565 & 0.266 & 0.358 & 0.331 & 0.190 \\ 
STS~\cite{wang2022sts} & R101 &\checkmark &\checkmark  & 0.431 & 0.542 & 0.525 & 0.262 & 0.380 & 0.369 & 0.204 \\
\midrule
\rowcolor[gray]{.9} 
\methodname{} & R101$^\dag$ & &  & \textbf{0.457} & \textbf{0.546} & \textbf{0.624}  &0.265 & 0.362 & \textbf{0.380} & 0.191 \\ 
\bottomrule
\end{tabular}}
\caption{\textbf{Comparison to state-of-art on the nuScenes \emph{val} set.} ${}^\dag$Initialized from FCOS3D~\cite{wang2021fcos3d} backbone. Our model is trained for 24 epochs without \emph{CBGS}~\cite{zhu2019class}. \emph{LiDAR} means training depth branch utilizing extra modality supervision from LiDAR.}
\vspace{-10pt}
\label{tab:main_val_set}
\end{table*}

\subsection{3D Object Detection Results}
We compare our method with the state of the art on both \emph{val} and \emph{test} sets of nuScenes.

\noindent \textbf{nuScenes test set.}
Table~\ref{tab:main_test_set} presents the results of our model on the nuScenes \emph{test} set, where we achieved a remarkable performance of \textbf{51.6 mAP} and \textbf{58.9 NDS} without utilizing any extra depth supervision from LiDAR.
Under the setting without utilizing LiDAR as supervision, our method outperforms the previous state of the art.
We evaluate our model in two types of backbone mentioned in the implementation details.
With R101-DCN~\cite{dai2017deformable} as the backbone, we could achieve  \textbf{47.8 mAP} and \textbf{56.1 NDS}, a significant improvement (+2.1 mAP and +1.8 NDS) over previous methods.
For the final performance, we train \methodname{} on the \emph{trainval} split for 24 epochs without CBGS~\cite{zhu2019class}, using VoVNet (V2-99) as the backbone architecture with a pre-trained checkpoint from DD3D~\cite{park2021pseudo}.

\noindent \textbf{nuScenes validation set.}
Table~\ref{tab:main_val_set} shows that our method achieves leading performance on the nuScenes \emph{val} set.
We achieved \textbf{45.7 mAP} and \textbf{54.6 NDS} without bells and whistles. Unlike the evaluation on \emph{test} set, all the methods are compared with a fair backbone here. 
Since BEVDepth~\cite{li2022bevdepth} and STS~\cite{wang2022sts} utilized extra modality supervision in training, our NDS metric only improved slightly compared to them, but our mAP improved significantly.
The translation error would be reduced with LiDAR supervision for the depth estimation, but this required extra modality data from LiDAR.
Besides, our model is trained for 24 epochs, while they actually trained 90 epochs if using CBGS~\cite{zhu2019class}.

\subsection{Ablation Study}
We conduct several ablation experiments on the nuScenes \emph{val} set to validate the design of \methodname{}.
For all ablation experiments, we used ResNet-50 as the backbone and resized the image resolution to 0.8 scales.

\noindent \textbf{Ablation of Components in \methodname{}.}
Table~\ref{tab:ablation} ablates the components designed in \methodname{}. 
(a) is the baseline setting of our method. 
(b) is the baseline with the instance frustum queries, which resamples the points in the whole instance frustum region. 
(c) is the baseline with the occupancy mask prediction. 
(d) is the baseline with adaptive instance-aware resampling, which consists of instance frustum query and occupancy mask prediction. Utilizing adaptive instance-aware resampling to improve the learning of instance-aware BEV feature can significantly enhance both mAP and NDS metrics.
(e) is based on (d) and further adds the history frustum information to incorporate temporal clues.
Above all, our \methodname{} could improve 4.2 mAP and 9.7 NDS compared to the baseline.
\begin{table}[ht]
\small
\centering
\begin{tabular}{c|ccc|ccc} 
\toprule
 & \textbf{IF} & \textbf{OM} & \textbf{FF} & \textbf{mAP}$\uparrow$  &\textbf{NDS}$\uparrow$  & \textbf{mATE}$\downarrow$  \\
\toprule
(a) &  &  & &0.318  & 0.366 & 0.771 \\
(b) & \checkmark & & &0.326  & 0.373 &0.765  \\
(c) &  &\checkmark &  &0.328 & 0.381 & 0.759 \\
(d) & \checkmark&\checkmark & & 0.337 & 0.383 & 0.749 \\
(e) & \checkmark & \checkmark & \checkmark  &\textbf{0.360}  & \textbf{0.463} & \textbf{0.719}\\
\bottomrule
\end{tabular}
\caption{\textbf{Ablation of components in \methodname{}.} IF denotes instance frustum, OM denotes occupancy mask, and FF means temporal frustum fusion. Adaptive instance-aware resampling is the combination of IF and OM, shown in (d).}
\label{tab:ablation}
\vspace{-10pt}
\end{table}

\noindent \textbf{Ablation of Instance-aware Sampling.}
Table~\ref{tab:ir} proves the effectiveness of instance-aware sampling.
(a) represents the baseline setting, which treats all regions equally and samples 1x points for a grid cell region.
(b) increases the sampling points to 2x for all regions. 
(c) selectively resamples the points inside instance regions.
Compared with (b) and (c), we found that instance-aware sampling is more effective since simply increasing the sampling points for all regions has no gain.
\begin{table}[ht]
    \small
    \centering
    \begin{tabular}{c|c|c|c|cc}
    \toprule
    &\textbf{Total} &\textbf{Scene} &\textbf{Instance}& \textbf{mAP}$\uparrow$ & \textbf{NDS}$\uparrow$ \\
    \toprule
    (a) & 1$\times$ & 1$\times$ &-& 0.318 & 0.366\\
    (b) & 2$\times$ & 2$\times$&- & 0.318  & 0.362 \\
    (c) & 2$\times$ & 1$\times$ & 1$\times$ & \textbf{0.326}   & \textbf{0.373} \\
    \bottomrule
    \end{tabular}
    \caption{\textbf{Ablation of instance-aware sampling.} In our method, 1$\times$ means sampling 8 points for a cell region on the BEV plane.}
    \vspace{-10pt}
    \label{tab:ir}
\end{table}

\noindent \textbf{Ablation of Occupancy Mask Learning.}
Table~\ref{tab:sampling} compares different supervision for learning occupancy masks on the BEV plane.
(a) is the baseline without explicit supervision. 
(b) adds the supervision on the BEV plane, in which the instance area can be obtained by projecting annotated bounding boxes. To ease the learning, we slightly enlarge the bounding boxes by 1.0 meters in this setting.
(c) uses the strict projection area (without enlargement) from the bounding boxes.
(d) increases the loss weight for $\mathcal{L}_m$ from 5.0 to 10.0.
(e) decreases the loss weight for $\mathcal{L}_m$ from 5.0 to 1.0.
We choose the (c) as our default setting.
\begin{table}[ht]
    \centering
    \small
    \begin{tabular}{c|c|c|ccc}
    \toprule
    &\textbf{Supervision} & \bm{$\alpha$}& \textbf{mAP}$\uparrow$ & \textbf{NDS}$\uparrow$ & \textbf{mATE}$\downarrow$ \\
    \toprule
    (a)& w/o & 0.0 & 0.318 & 0.366 & 0.771 \\
    (b) & w/ BEV box* &5.0& 0.324  & 0.374 & 0.756 \\
    (c)& w/ BEV box &5.0 & \textbf{0.328}  & \textbf{0.381}  & 0.759 \\
    (d)& w/ BEV box &10.0 & 0.322  & 0.381  & \textbf{0.749} \\
    (e)& w/ BEV box &1.0 & 0.326  & 0.379  & 0.751 \\
    \bottomrule
    \end{tabular}
    \caption{\textbf{Ablation of occupancy mask learning.} BEV box means utilizing the ground truth bounding boxes' projection on the BEV plane as supervision. * denotes enlarged bounding box. $\alpha$ is the loss weight for learning occupancy mask.}
    \label{tab:sampling}
    \vspace{-10pt}
\end{table}

\noindent \textbf{Ablation for Temporal Frustum Fusion.}
In Table~\ref{tab:time_window}, we demonstrate the benefits of incorporating frustum information in temporal fusion, as well as examine the impact of two important parameters: window size $W$ and keyframes $K$.
We start with the baseline (a) that uses a window size of 4 and 2 keyframes. Subsequently, we introduce the temporal frustum information in (b), leading to a notable improvement in performance. Then, we experiment with a larger temporal window by setting $W$ to 8 and $K$ to 4 in (c), which yields the best performance. As we extend the temporal window further in (d), the mAP continues to improve, but the NDS metric is affected by the mAVE score, causing degradation.
Considering the NDS and mAP comprehensively, we finally utilize setting (c) as the default.
\begin{table}[ht]
    \small
    \centering
    \begin{tabular}{c|c|c|c|cc|c}
    \toprule
    & \textbf{W}   &  \textbf{K} & \textbf{Frustum} & \textbf{mAP}$\uparrow$ & \textbf{NDS}$\uparrow$ & \textbf{mAVE}$\downarrow$\\
    \toprule
        (a) & 4 & 2& &0.353& 0.454&0.497\\
        (b) &4 & 2& \checkmark& 0.355 & 0.457&0.479 \\
        (c) &8 & 4& \checkmark &0.360 & \textbf{0.463} & \textbf{0.463} \\
        (d) & 16 & 4 & \checkmark& \textbf{0.364} & 0.457 & 0.568 \\
    \bottomrule
    \end{tabular}
    \caption{\textbf{Ablation for temporal frustum fusion.} W means the history window size. K determines the key frames sampled in temporal window during model training.}
    \label{tab:time_window}
    \vspace{-10pt}
\end{table}

\begin{figure}[ht]
    \includegraphics[width=1.0\linewidth]{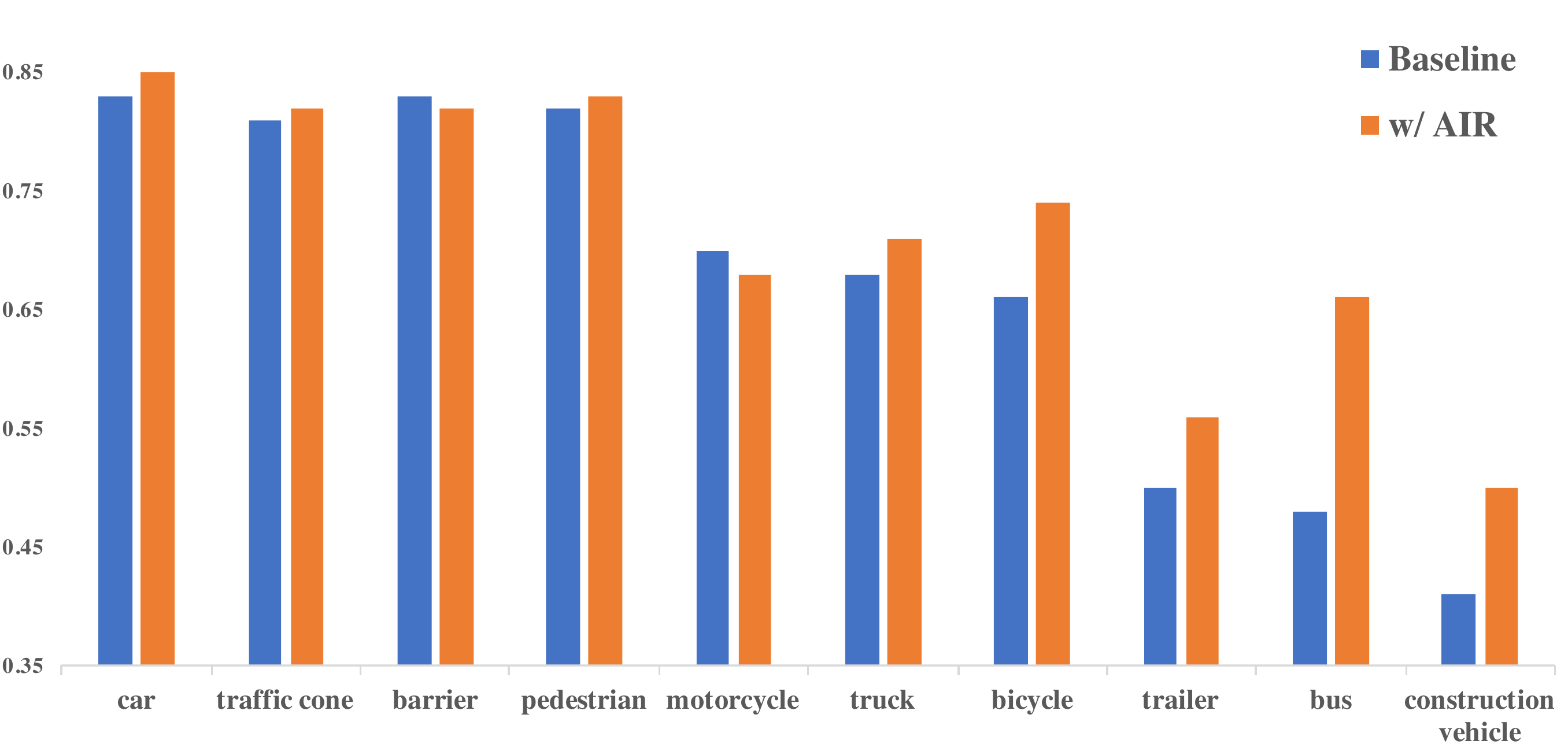}
    \caption{\textbf{Improvement of Recall Under Low Visibility.} We compute the recall under the visibility at 0-40\% for all categories on the nuScenes \emph{val} set. The recall for bus, bicycle, trailer, and construction vehicle categories improved significantly.}
    \label{fig:recall}
    \vspace{-10pt}
\end{figure}

\noindent \textbf{Improvement of Recall Under Low Visibility.}
The nuScenes dataset provides the visibility labels of objects in four subsets $\{$0-40\%, 40-60\%, 60-80\%, 80-100\%$\}$. 
As shown in \cref{fig:recall}, we compare the recall between baseline and baseline with adaptive instance-aware resampling under low visibility (0-40\%).
We found that the recall for categories of \emph{bicycle}, \emph{trailer}, \emph{bus}, and \emph{construction vehicle} improved a lot under the low visibility.
Since nearly 29\% of objects belong to the visibility of 0-40\%, such improvement is crucial for a better 3D object detector.

\subsection{Qualitative Analysis}

\begin{figure}[ht]
    \includegraphics[width=1.0\linewidth]{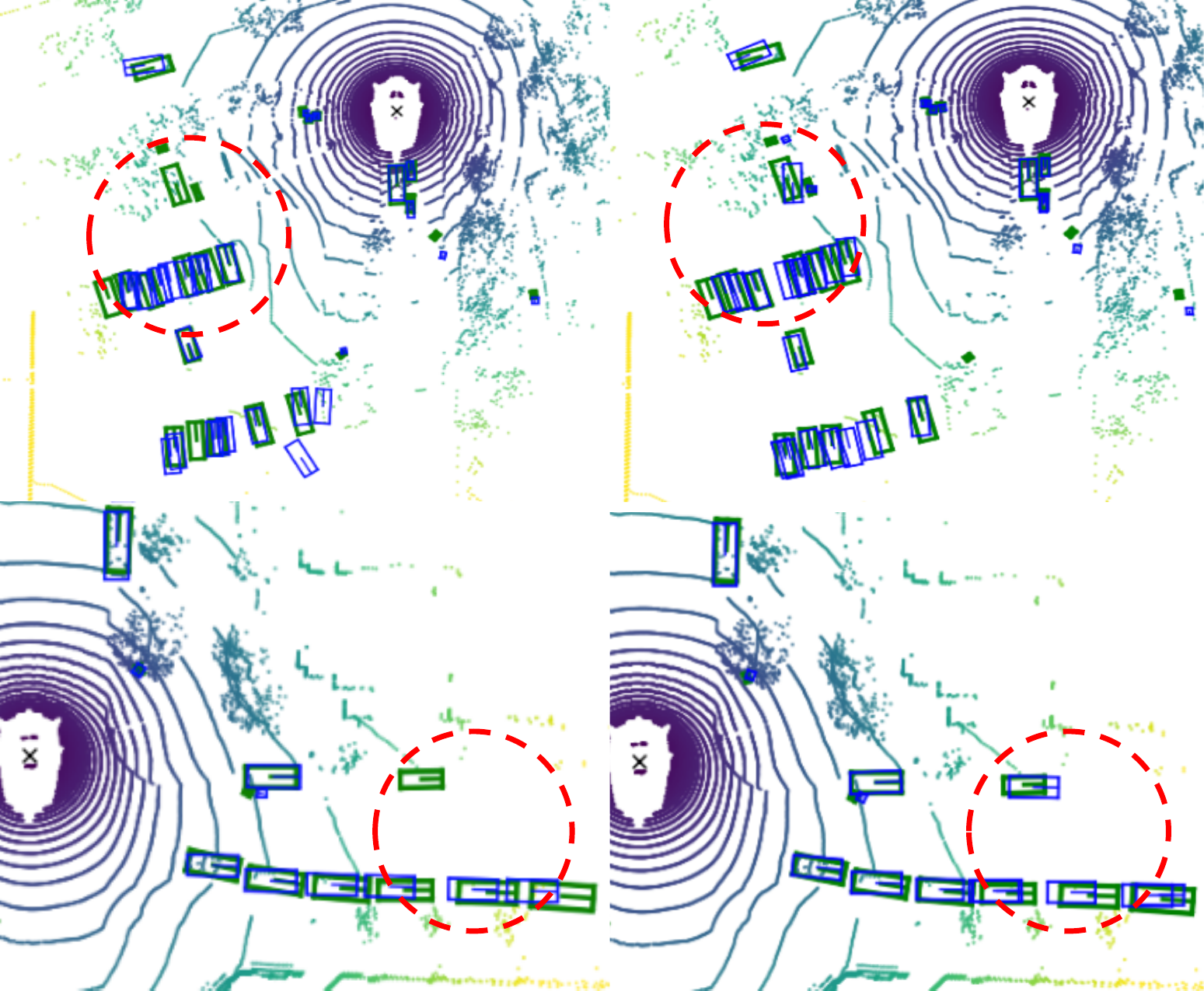}
    \caption{\textbf{Visualization of Recall Improvement}. The left side shows the baseline, while the right side shows the baseline with AIR. The prediction boxes are marked in blue, while the ground truth boxes are marked in green. In the red circle region, our method discovers more objects than the baseline.}
    \label{fig:convergence}
    \vspace{-10pt}
\end{figure}

\noindent \textbf{Visualization of Recall Improvement.}
As shown in \cref{fig:convergence}, we present the recall improvement achieved by utilizing the adaptive instance resampling (AIR) approach on the prediction results of the nuScenes \emph{val} dataset. The predicted boxes are highlighted in blue, while the ground truth boxes are displayed in green. The red circle indicates the region where AIR helps to discover objects that were previously missed by enhancing the learning of instance features.

\begin{figure}[ht]
  \centering
    \includegraphics[width=1.0\linewidth]{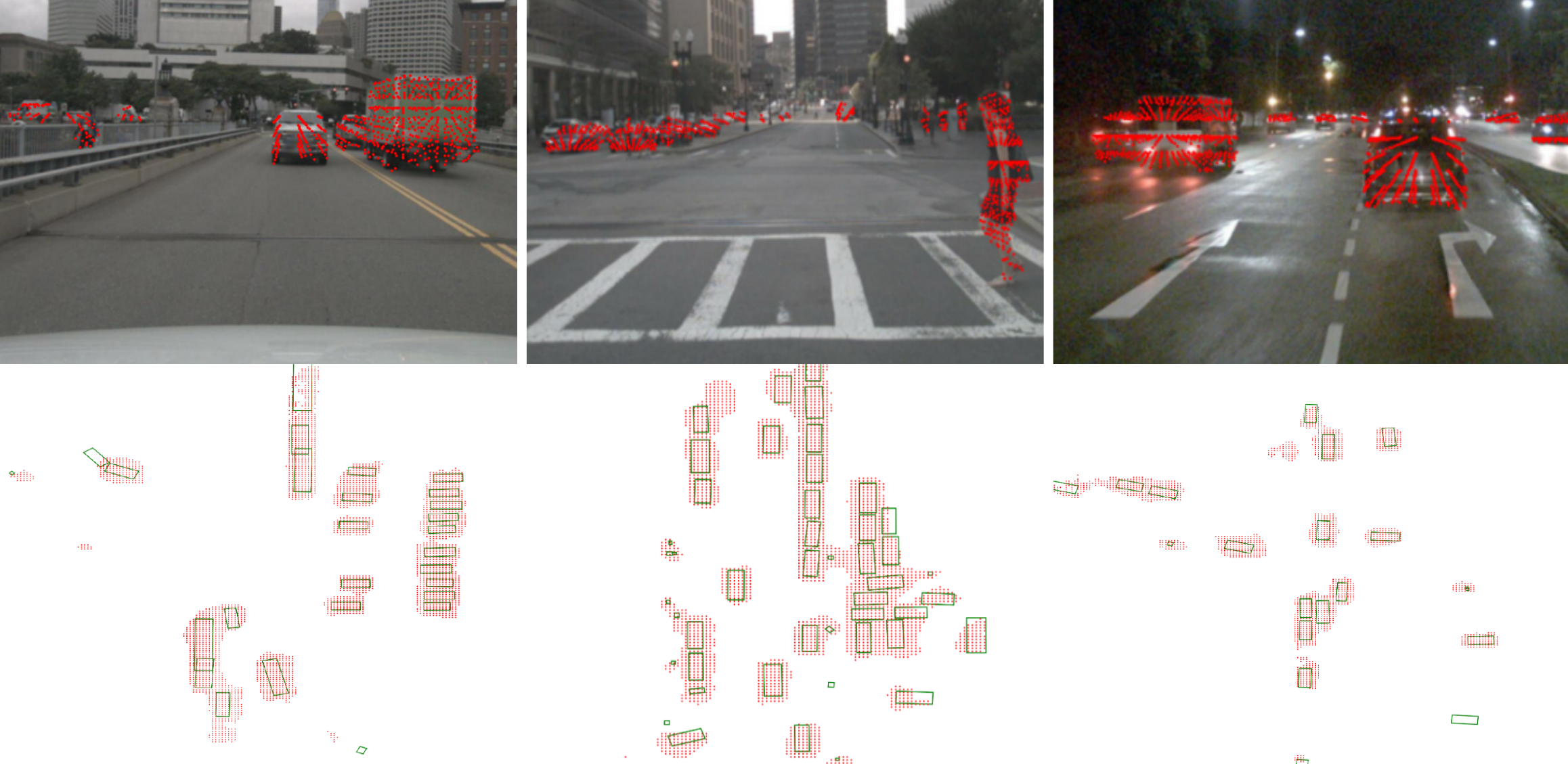}
    \caption{\textbf{Visualization of Instance-aware Sampling.} We visualize the instance-aware sampling on the perspective view and bird's eye view. Ground truth bounding boxes are marked in green color.}
  \label{fig:sampling}
   \vspace{-10pt}
\end{figure}

\noindent \textbf{Visualization of Instance-aware Sampling.}
As shown in \cref{fig:sampling}, we illustrate the instance-aware sampling points of our method on both perspective view and bird's eye view. The sampling points are highly related to the instance regions, enhancing the learning of the instance-aware feature.

\begin{figure}[ht]
    \vspace{-5pt}
     \includegraphics[width=1.0\linewidth]{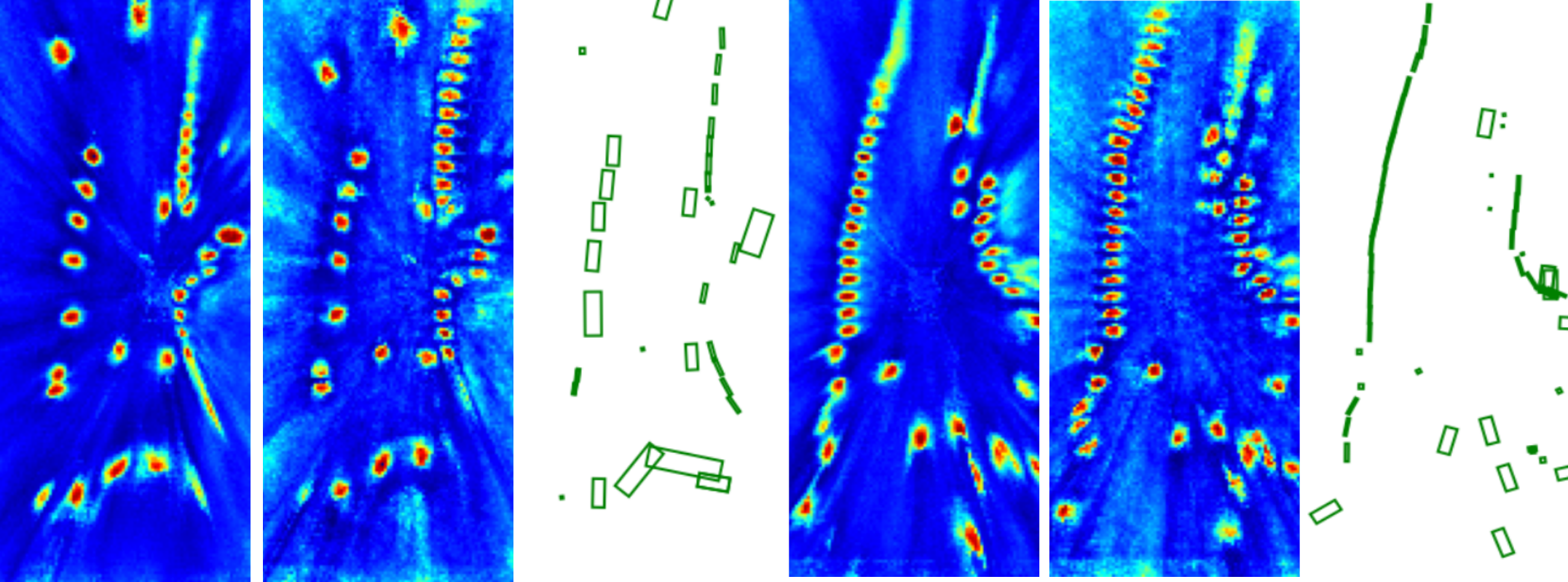}
    \caption{\textbf{Visualization of the instance-aware BEV feature.} We compared the feature heatmaps output by the frustum encoder without AIR, with AIR, and ground truth boxes (in green) shown from left to right. The colors in the feature heatmaps correspond to the norm value. The feature learned by frustum encoder is instance-aware and has strong correlations to the actual positions in 3D space. By using AIR, our model is able to discover more instance regions and learn discriminative features in dense areas.}
    \label{fig:bevfeature}
    \vspace{-10pt}
\end{figure}

\noindent \textbf{Visualization of Instance-aware BEV Feature.}
As shown in ~\cref{fig:bevfeature}, we visualize the BEV feature output by our frustum encoder. The BEV feature learned by our frustum encoder is instance-aware and has strong relations to the real positions. Here we visualize the corresponding ground truth boxes on the right side.
Furthermore, we compare the BEV feature between the baseline and the baseline with adaptive instance-aware resampling(AIR). When utilizing AIR, more instance regions would be discovered (corresponding to the recall improvement), and the features are more instance discriminative in the dense areas.

\section{Conclusion}

In this paper, we propose \emph{\methodname{}}, a novel framework for multi-view 3D object detection.
The core insight of \methodname{} is to transform adaptively according to the view contents.
To achieve this, we designed adaptive instance-aware resampling, which pays more attention to instance regions during feature view transformation. By utilizing this technique in the frustum encoder and temporal frustum fusion module, the model learned to better locate instance regions while learning instance-aware BEV features.
Experimental results on the nuScenes dataset demonstrate the effectiveness of our method for multi-view 3D object detection. 
Our method significantly improved mAP over previous methods by focusing on instance regions.
We hope that our framework can serve as a new baseline for future 3D perception research, shining a light on the significance of view content during feature transformation.
\section*{Acknowledgments}
This work was supported in part by the Major Project for New Generation of AI (No.2018AAA0100400), the National Natural Science Foundation of China (No.61836014, No.U21B2042, No.62072457, No.62006231) and the InnoHK program. We extend our sincere thanks and appreciation to Jiawei He and Lue Fan for their valuable suggestions.
{\small
\bibliographystyle{ieee_fullname}
\bibliography{egbib}
}

\end{document}